\documentclass[conference]{IEEEtran}
\IEEEoverridecommandlockouts
\usepackage{cite}
\usepackage{amsmath,amssymb,amsfonts}
\usepackage{algorithmic}
\usepackage{graphicx}
\usepackage{textcomp}
\usepackage{xcolor}
\usepackage{caption}
\usepackage{subcaption}
\def\BibTeX{{\rm B\kern-.05em{\sc i\kern-.025em b}\kern-.08em
    T\kern-.1667em\lower.7ex\hbox{E}\kern-.125emX}}
\begin{document}






\title{Evaluating the acceptance of autonomous vehicles in the future\\

\thanks{This work has been supported by the Spanish Government through the projects ID2021-128327OA-I00 and TED2021-129374A-I00 funded by MCIN/AEI /10.13039/501100011033 and by the European Union NextGenerationEU/PRTR  and   Madrid Government (Comunidad de Madrid-Spain) under the Multiannual Agreement with UC3M (“Fostering Young Doctors Research”, APBI-CM-UC3M), and in the context of the V PRICIT (Research and Technological Innovation Regional Programme) also through MAPFRE through CESVIMAP.}
}

\author{

\IEEEauthorblockN{Angel Madridano Carrasco }
\IEEEauthorblockA{\textit{Systems Eng. and Automatics Dep.} \\
\textit{Universidad Carlos III de Madrid}\\
Madrid, Spain \\
amadrida@ing.uc3m.es }
\and
\IEEEauthorblockN{ Delgermaa Gankhuyag}
\IEEEauthorblockA{\textit{Chair for Sustainable Transport
Logistics 4.0,} \\
\textit{Johannes Kepler University}\\
Linz, Austria \\
delgermaa.gankhuyag@jku.at }
\and
\IEEEauthorblockN{Miguel Angel de Miguel Paraiso }
\IEEEauthorblockA{\textit{Systems Eng. and Automatics Dep.} \\
\textit{Universidad Carlos III de Madrid}\\
Madrid, Spain \\
mimiguel@ing.uc3m.es }
\and

\IEEEauthorblockN{Martin Palos Lorite }
\IEEEauthorblockA{\textit{Systems Eng. and Automatics Dep.} \\
\textit{Universidad Carlos III de Madrid}\\
Madrid, Spain \\
mpalos@ing.uc3m.es }
\and
\IEEEauthorblockN{ Cristina Olaverri-Monreal}
\IEEEauthorblockA{\textit{Chair for Sustainable Transport
Logistics 4.0,} \\
\textit{Johannes Kepler University}\\
Linz, Austria \\
cristina.olaverri-monreal@jku.at }
\and
\IEEEauthorblockN{ Fernando Garcia Fernandez}
\IEEEauthorblockA{\textit{Systems Eng. and Automatics Dep.} \\
\textit{Universidad Carlos III de Madrid}\\
Madrid, Spain \\
fegarcia@ing.uc3m.es }

}

\maketitle

\begin{abstract}
The continuous advance of the automotive industry is leading to the emergence of more advanced driver assistance systems that enable the automation of certain tasks and that are undoubtedly aimed at achieving vehicles in which the driving task can be completely delegated. All these advances will bring changes in the paradigm of the automotive market, as is the case of insurance. For this reason, CESVIMAP and the Universidad Carlos III de Madrid are working on an Autonomous Testing pLatform for insurAnce reSearch (ATLAS) to study this technology and obtain first-hand knowledge about the responsibilities of each of the agents involved in the development of the vehicles of the future. This work gathers part of the advancements made in ATLAS, which have made it possible to have an autonomous vehicle with which to perform tests in real environments and demonstrations bringing the vehicle closer to future users. As a result of this work, and in collaboration with the Johannes Kepler University Linz, the impact, degree of acceptance and confidence of users in autonomous vehicles has been studied once they have taken a trip on board a fully autonomous vehicle such as ATLAS. This study has found that, while most users would be willing to use an autonomous vehicle, the same users are concerned about the use of this type of technology. Thus, understanding the reasons for this concern can help define the future of autonomous cars.

\end{abstract}

\begin{IEEEkeywords}
 Autonomous driving, Intelligent Transportation Systems, Survey, Social impact.
\end{IEEEkeywords}

\section{Introduction}
\label{sec: Introduction}
One of the greatest challenges facing today's society is to achieve safe, efficient and sustainable mobility for both goods and people~\cite{b1, b2}. To meet this challenge, the automotive industry and the scientific community are focusing their efforts on developing technology that will make it possible to have vehicles whose driving is fully automated, i.e., means of ground transportation with level 5 automation, as established by the Society of Automotive Engineers (SAE)~\cite{b3}. 

Currently, road traffic fatalities result in around 1.3 million deaths per year~\cite{b4}. In addition, according to data from the World Health Organization, road traffic injuries are the leading cause of death among children and young people aged 5-29 years~\cite{b7, b5}. However, this negative impact is not only reflected in terms of safety; according to European Union data, transport continues to have a high negative impact on the environment, being the sector that emits the highest percentage of greenhouse gases into the atmosphere, with 28\%~\cite{b6}. 

These data are a brief summary of why the competent authorities and agents in the sector are currently working together on this mobility challenge. As a result of this effort and dedication, great progress has been made in the field of Autonomous Vehicles (AVs). Among the most outstanding examples are the developments made by the manufacturer Audi, with a vehicle capable of traveling 900 $Km$ autonomously in USA~\cite{b7}; the fleet of autonomous vehicles of the company Google, which is currently circulating without supervision in a delimited area of Phoenix and which accumulates more than 2 million kilometers traveled completely autonomously~\cite{b8, b9}; In addition, within these advances, the 'Drive Pilot' of the manufacturer Mercedes~\cite{b10}, which is included in commercial models since 2022, allowing drivers to delegate driving to the vehicle in certain conditions, or Tesla's 'Autopilot', both systems that provide autonomy to the vehicle~\cite{b11}, but are far from the highest SAE levels.  

Despite all the successes achieved in recent years in the field of autonomous driving, the fact of having completely autonomous vehicles in cities seems to be very far away, and it is not only technology that must continue to advance to ensure high levels of safety, but also legislation and society must adapt to the emergence of highly sophisticated systems that will drastically change the way people move around the world. The adaptation of people to these new technologies, together with the social and cultural consequences derived from the use of AVs, is a growing field of research. Among the aspects to be addressed within this topic is how the irruption of AVs will affect public transportation systems, which will generate a conflict of interest by allowing people without a driver's license to move around more comfortably and directly~\cite{b12}.

Another important issue is to know the degree of trust people have in this technology. Therefore, works such as~\cite{b13} or~\cite{b14} collect studies about this issue, focusing on understanding the actions and reactions of users to these vehicles and conducting surveys and analysis of the degree of trust in AVs. The results of these studies are beneficial to help developers and manufacturers to optimize this new technology by improving its adoption in society. Other examples of these studies are Volvo's "DriveMe" projects, in which users could interact with the AVs and share their experiences in order to improve the studies of interaction between humans and such technology~\cite{b15}. In line with these studies,~\cite{b16}reports the analysis of an experiment conducted in Austria in which the response of pedestrians when crossing a street while an AV was approaching was studied.

For these reasons, CESVIMAP, the R\&D division of the MAPFRE automobile insurance company, has decided to promote a project in which, starting from a mass-produced vehicle, an automation process is carried out to achieve a research platform capable of reaching SAE level 4 automation. Thus, ATLAS is a prototype developed jointly by CESVIMAP, the Universidad Carlos III de Madrid and the Universidad Politecnica de Madrid, with which the insurer seeks to understand all the key elements and the necessary steps required to bring a vehicle of these characteristics into circulation and, to delimit the competences and responsibilities of the manufacturers, the suppliers of sensors and equipment, the software developers, the vehicle owner and the occupants of the vehicle.

This article includes in section~\ref{secII: AtlasProject} part of the technological improvements made by CESVIMAP and Universidad Carlos III de Madrid, which have made it possible to carry out real-world experiments with the ATLAS platform circulating autonomously in previously defined and delimited environments and allowing people to travel aboard the vehicle and see for themselves what the experience of being driven by a driverless vehicle is like. Thanks to these experiments, and in collaboration with the Johannes Kepler University Linz, it has been possible to study the impact of this technology in real users, the degree of acceptance of this type of vehicle and the predisposition of people to use driverless vehicles, as described in section~\ref{secIII:Results}. Finally, section~\ref{secIV:Conclusions} summarizes the conclusions of the work and establishes the tasks and steps to be undertaken in the future.

\section{Research platform and public experiments}
\label{secII: AtlasProject}

This section describes the most relevant characteristics of the ATLAS platform and the main aspects of the test environment in which the autonomous driving experiments gathered in this work were conducted. 

\subsection{ATLAS Platform}
\label{susec: ATLASPlatform}

\begin{figure}[htbp]
\centerline{\includegraphics[width=0.75\linewidth]{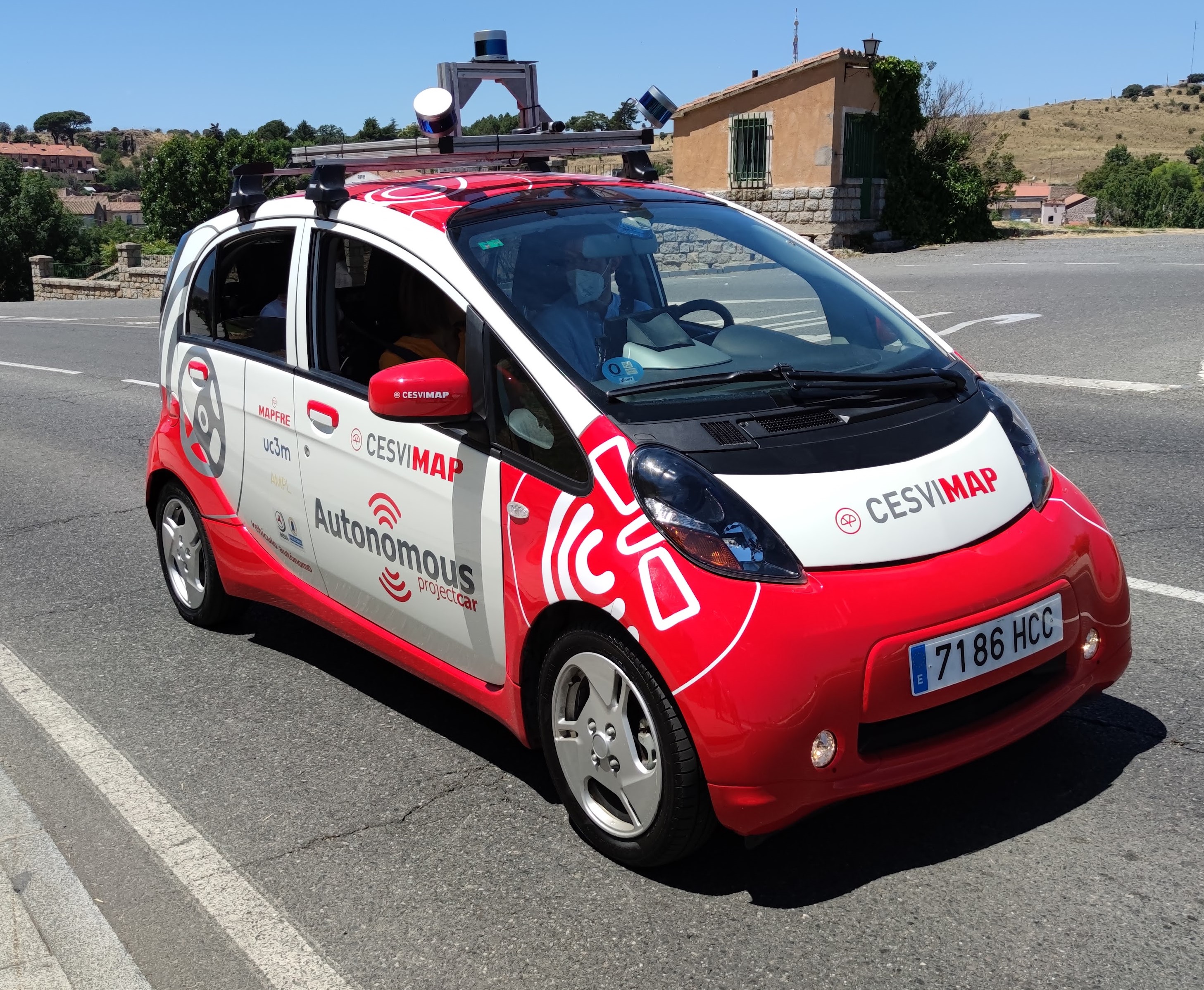}}
\caption{ATLAS Platform.}
\label{fig:Atlas}
\end{figure}

The ATLAS research platform is, at its base, a Mitsubishi iMiev vehicle as shown in Figure~\ref{fig:Atlas}. The main objective of this system is to advance in the development and testing of onboard autonomous vehicle technologies and to allow companies such as MAPFRE to obtain an in-depth knowledge of these systems, in order to adapt their insurance services to the mobility of the future. This platform makes it possible to test and learn about different types of sensors, technologies and implementations, enabling the acquisition of an exhaustive knowledge regarding each of the agents involved in the manufacture and development of these vehicles. 

Thus, ATLAS gives name to a project that allows to focus scientific work in the area of ITS, which encompasses a complete view of all its aspects, unlike other research in the area of autonomous driving more oriented on the objective evaluation of specific components and systems of the vehicle. This aspect has allowed ATLAS to be used to understand, analyze and evaluate the impact of this technology on society through a heterogeneous group of people, all of whom have one element in common, which is having been able to complete a predefined route in a fully autonomous vehicle. 

As previously mentioned, ATLAS is established as a SAE level 4 autonomous vehicle. To reach this level of automation and, to transform a commercial car into an intelligent platform, several years of research and development have been necessary since 2018. Although it is true that, throughout these years, the continuous advances and the evolution of technology have led to the introduction of new systems and components in the vehicle, most of the characteristics and developments of the vehicle are included in the work~\cite{b20}. While a brief description of the current state of the vehicle will be provided below, \cite{b20} offers an in-depth explanation of aspects of the platform that are not covered in this section. 

\begin{figure}[htbp]
\centerline{\includegraphics[width=0.98\linewidth]{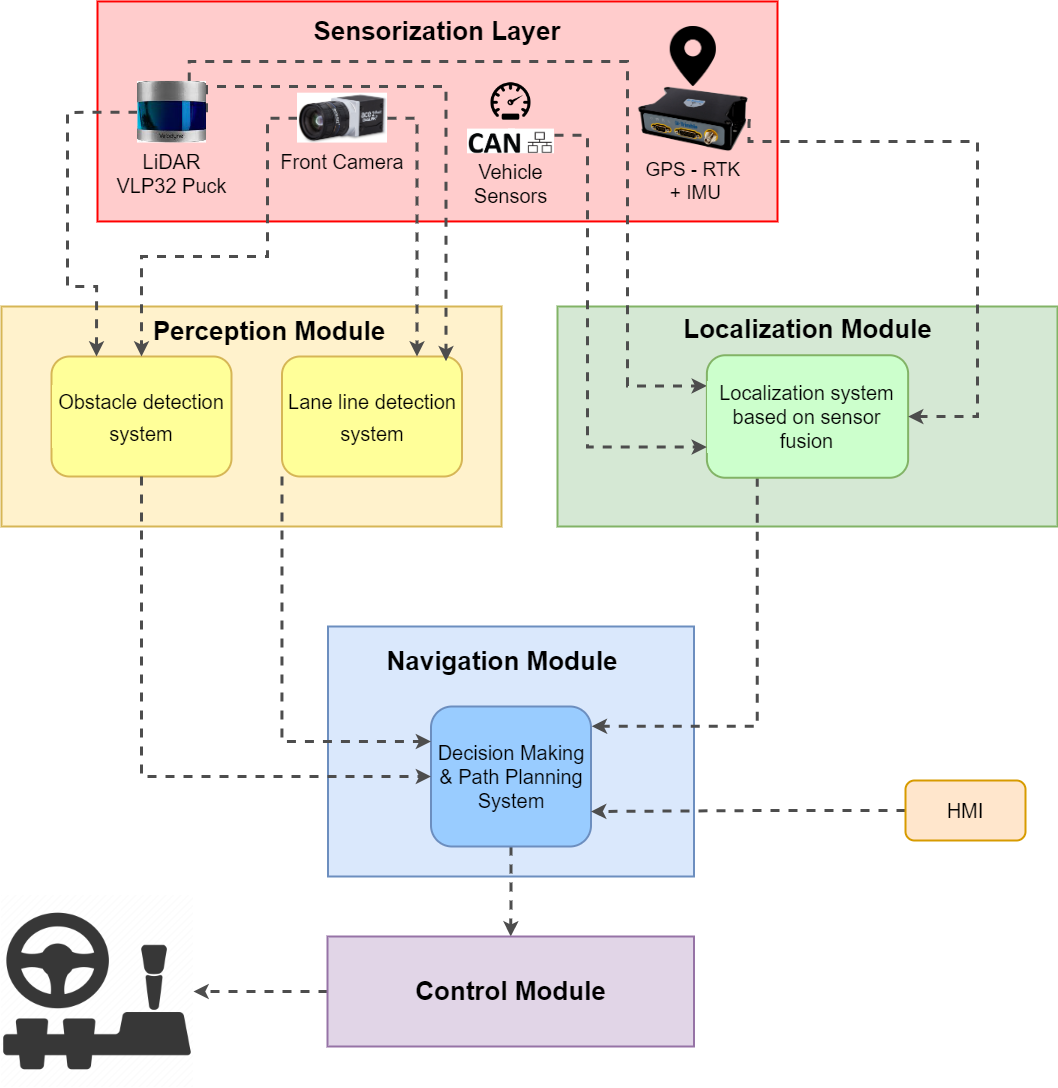}}
\caption{ATLAS Architecture.}
\label{fig:AtlasArchitecture}
\end{figure}

Currently, ATLAS has a Hardware / Software architecture shown and summarized in Figure~\ref{fig:AtlasArchitecture}. As it can be observed, the vehicle is equipped with a set of sensors in charge of acquiring all the necessary information from the environment and the vehicle itself. This sensors layer comprises a GNSS localization system with RTK correction complemented with an inertial sensor or IMU, a perception system consisting of a 32-layers LiDAR and a front monocular camera and, finally, the vehicle's own set of sensors that provide data such as the vehicle speed or the steering wheel turning angle. All the data collected by the sensors is interpreted by the software architecture, which is composed by different modules whose synchronized and coordinated operation allows the vehicle to act autonomously and safely. These modules are:

\begin{enumerate}
    \item \textbf{Perception Module.}
    It processes the 2D and 3D information collected by the camera and the main LiDAR respectively and generates an output accordingly in order to assist the navigation module in its decision-making. It is important to point out that the sensors extrinsic parameters are calibrated so that all algorithms implemented can fuse the information coming from both sensors. This module is divided into two methods.
    \begin{itemize}
        \item \textbf{\textit{Line lane detection system:}} It is one of the car's main systems. It uses the information received from the camera and LiDAR to detect the lane in which the vehicle is and provide the control module with the information required to keep the vehicle in the centre of its lane, provided that the detection is correct. Using Deep Learning, a road line detection is performed on the camera-generated images, then this information is merged with the LiDAR three-dimensional data, in order to accurately determine the lines of the road through regressions.  
        
        \item \textbf{\textit{Obstacle detection system:}} Based on advanced Deep Learning techniques, it uses Neural Networks to detect, classify and locate obstacles in the environment with respect to the vehicle. For this system, and due to computational limitations, only three-dimensional LiDAR information is used as input. 
    \end{itemize}
    \item \textbf{Localization Module.} It is responsible for accurately determining the position and orientation of the vehicle in the world. This module has a GPS receiver with RTK corrections as its core, whose information is merged with other data sources such as the vehicle's odometry or the LiDAR. The localization module is responsible for providing the platform with a real-time centimetre-accurate location, required for the correct and safe navigation of the vehicle in the environment. 
    
    \item \textbf{Navigation Module.} It manages the centralisation of most of the information generated by the aforementioned components of the vehicle to establish decision-making and path planning strategies to ensure safe vehicle navigation. In addition, as shown in Figure~\ref{fig:AtlasArchitecture}, the interaction of the vehicle with the user through a Human-Machine Interface (HMI) is used as an additional input. This interface makes it possible to act on the vehicle to change the operating mode between manually-driven and autonomous and to monitor the status of the platform and on-board systems.
    
    Furthermore, this module includes all the software in charge of path planning and, therefore, it is in this module where the user establishes the path to follow, setting a set of waypoints to navigate through, the speeds to traverse through these waypoints, possible mandatory stop points and, most importantly, it defines the regions in which the lane detection system is not accurate (roundabouts, areas without road lane delimitation) so that the navigation system relies solely on the vehicle's localization system. In this way, the vehicle has two complementary navigation methods, one based on the detection of road lines and, consequently, the positioning of the vehicle in the centre of its lane, and a second method in which the location of the vehicle generated by the localization system is taken as the only reference. Finally, when navigating, computing trajectories and making decisions, the navigation module always considers the data processed by the obstacle detection system to ensure collision-free navigation. 

    \item \textbf{Control Module.} It generates control commands on the vehicle actuators. It processes the information generated by the navigation module and acts accordingly to perform safe actions on the vehicle control elements. This module is composed of control methods that translate high-level instructions (turning, braking, reaching a certain speed value) into low-level control commands (throttle actuation, brake actuation, desired angle of steering wheel). 
\end{enumerate}

\subsection{Test environment}
\label{susec: TestEnvironment}

\begin{figure}[htbp]
\centerline{\includegraphics[width=0.9\linewidth]{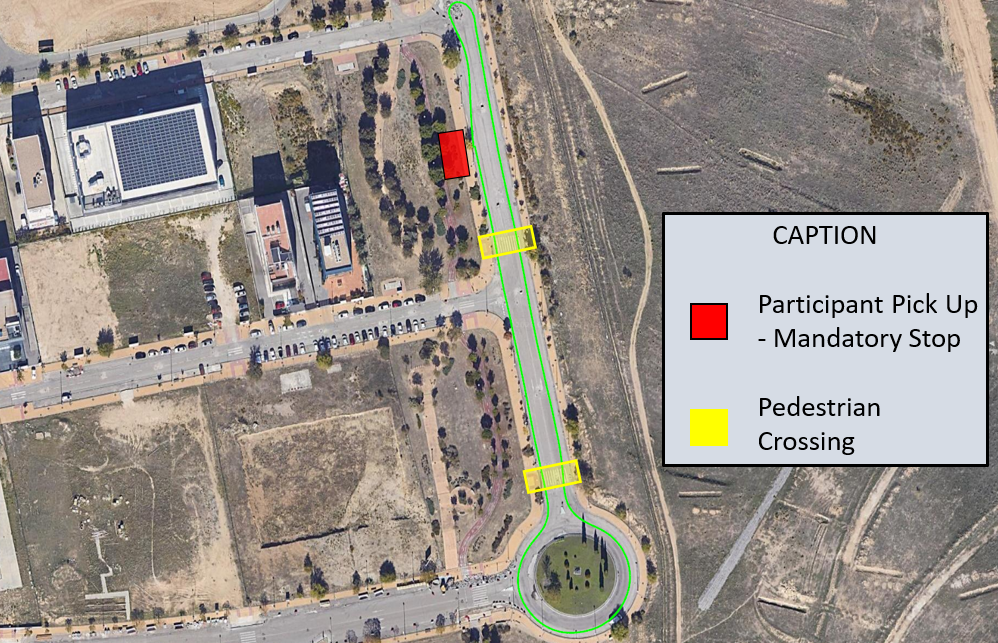}}
\caption{Test path (in green).}
\label{fig:TestEnvironment}
\end{figure}

After describing the ATLAS platform, this section will detail the test environment used to collect the survey data described and analyzed in the results section. The circuit used for the demonstration, in which survey participants have traveled inside the vehicle driving autonomously, is shown in green in Figure~\ref{fig:TestEnvironment}. This circuit, with a total length of 600 $m$, is located within the Technological Park of the city of Leganés (Spain) and, as it can be seen, is composed of several challenging road elements such as pedestrian crossings or roundabouts.

Figure~\ref{fig:TestEnvironment} depicts the place from where the demonstration starts (red square). This area of the path is set as a mandatory stop so that the autonomous vehicle can stop and passengers can get on and off the vehicle safely. Once passengers are in the vehicle, the platform can be instructed from the HMI to start driving. The demonstration consists of travelling 222 $meters$ in a straight line by detecting the road lane lines and evaluating both possible obstacles present on the road and pedestrians that may cross the crosswalk (yellow zones in Figure~\ref{fig:TestEnvironment}). Thus, if the vehicle detects people crossing the road at the crosswalks, it will stop until they have left the roadway and are out of the vehicle's detection zone. The rest of the path comprises a roundabout and another straight section, completing the aforementioned 600 meters. 

Therefore, the experience of the users interviewed in the survey whose results are collected in Section~\ref{secIII:Results} was as follows: the vehicle stopped autonomously in the pick up area; once the passengers were in the vehicle, the HMI was used to inform the platform that it could resume driving. From this moment on, the vehicle drove autonomously at a speed up to 30 $km/h$  (maximum speed allowed) along the predefined route, using the road line detection system as the main navigation method up to the roundabout section, where due to the curvature of the road, the algorithm is not able to perform a precise road line detection so the GPS-based localization is used to follow the predefined path. Once the roundabout is completed, the vehicle operation is again controlled by the line detection system. At all times, the obstacle detection system is operational, sending instructions to the control system if necessary. Due to current regulations, traffic was restricted during the experiment and, although vehicles were allowed to enter the area, it was done in such a way that their interaction with ATLAS was prevented (e.g. oncoming traffic when ATLAS is on the straight section). Even so, obstacle detection makes it possible to identify and locate other possible hazards such as pedestrians, causing the vehicle to stop in case of detecting an obstacle that interferes with the vehicle's navigation. Lastly, in the final section of the path, there is a sharp left turn to close the circuit, which, due to the need to cross the lanes perpendicularly, is performed using the GPS-based positioning. Once the turn has been completed, the vehicle reaches the pick up point, stops, passengers are switched, and the cycle is repeated again. 

On each lap, two passengers testing the vehicle for the first time were accompanied by an expert driver who supervised the correct operation of the system. During the demonstration, ATLAS travelled a total distance of approximately 9 $Km$ in a completely autonomous manner, without any incidents and acting correctly when detecting pedestrians crossing the road at the designated locations. At the end of the trip, the volunteer passengers completed the survey whose results are detailed and analyzed in the following section.

\section{Results}
\label{secIII:Results}

The survey performed contains information on people's interest in the automotive world, their knowledge in the field of autonomous vehicles, their experience with an autonomous driving car and their willingness to use that type of car. 

\begin{figure}[h]
     \centering
     \begin{subfigure}[b]{0.49\linewidth}
         \centering
         \includegraphics[width=\linewidth]{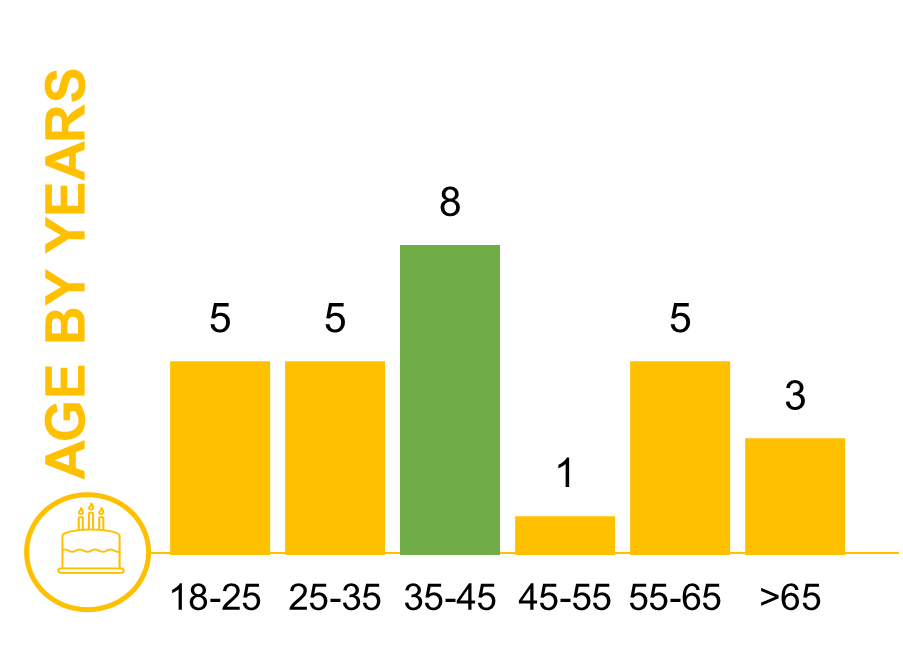}
         \label{fig:Ages}
     \end{subfigure}
     \begin{subfigure}[b]{0.49\linewidth}
         \centering
         \includegraphics[width=\linewidth]{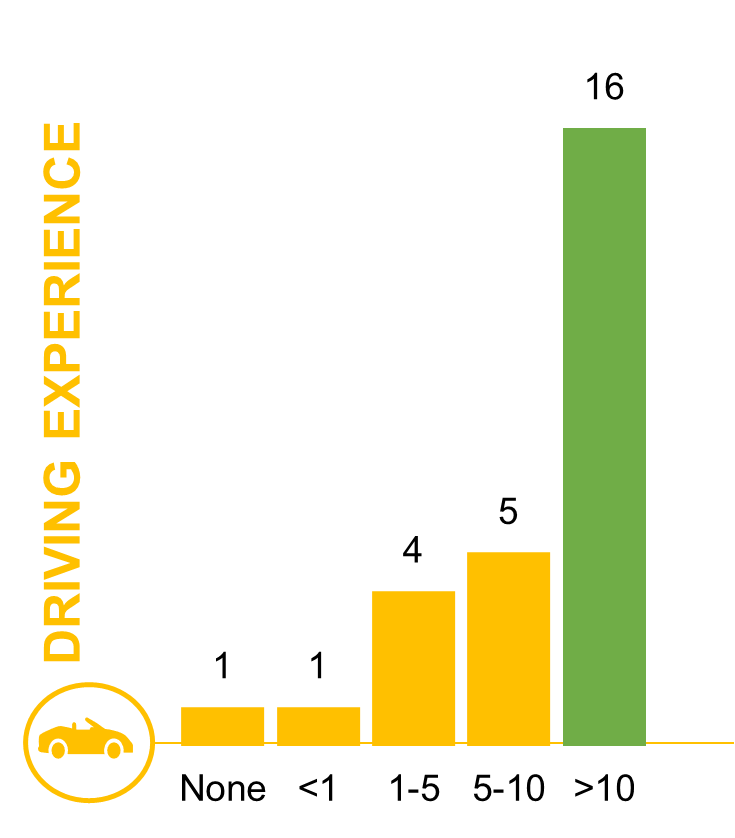}
         \label{fig:Experience}
     \end{subfigure}
     \begin{subfigure}[b]{0.49\linewidth}
         \centering
         \includegraphics[width=\linewidth]{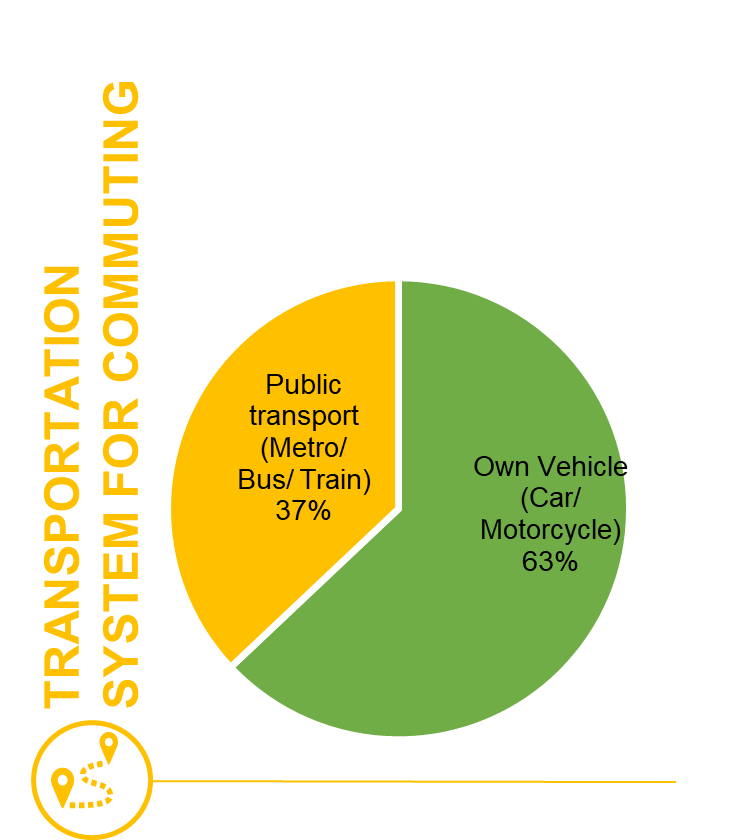}
         \label{fig:TransportSytem}
     \end{subfigure}
     \begin{subfigure}[b]{0.49\linewidth}
         \centering
         \includegraphics[width=\linewidth]{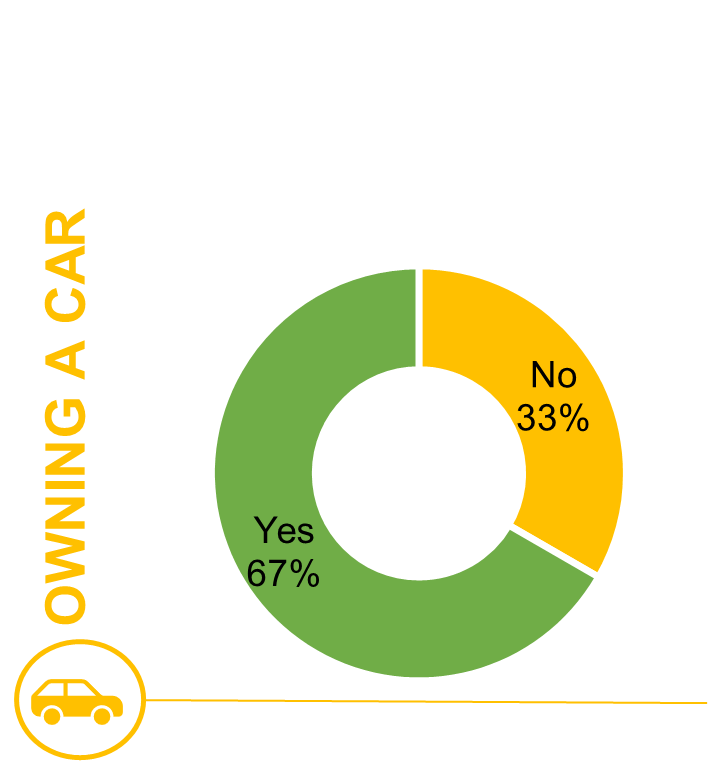}
         \label{fig:OwningCar}
     \end{subfigure}
        \caption{Demographic statistics.}
        \label{fig:Demographicsstatistics}
\end{figure}

It is relevant to note that all respondents had previously participated in the autonomous vehicle demonstration described above, so their input is therefore of interest, as it allows us to assess how knowledge and user experience in the area influences the trust and acceptance of this technology in society. Although a detailed analysis of various factors will be carried out throughout the section, a preliminary study of the survey's results reveals that almost all people are willing to use and interact with an autonomous vehicle and even encourage people whose opinion is important to them to use this type of vehicle. In addition, all people felt safe in the car and stated that they would reach their destination without any problems. This may seem surprisingly high and demonstrates the participants' great interest and confidence in the experiment's vehicle. 

A total of 27 people participated in the survey (14 females and 12 males). As shown in the Figure~\ref{fig:Demographicsstatistics}, a representative of all age segments was chosen for the survey. However, there was only one person from the 45-55 age segment, whereas eight people from 35-45 age. Most of those who participated had more than ten years of driving experience ($>$ 60\%).  

One-third of them had one to ten years of driving experience. Furthermore, a minority of candidates (2 people) had less than one year of driving experience or no driving license. Approximately two third of people (67 percent) owned a car, whereas one-third (33 percent) did not own one. Those not owning a car used public transportation such as the metro, bus, and train daily; on the other hand, almost all car owners used their vehicles for daily commuting.

\begin{figure}[h]
     \centering
     \begin{subfigure}[b]{0.98\linewidth}
         \centering
         \includegraphics[width=\linewidth]{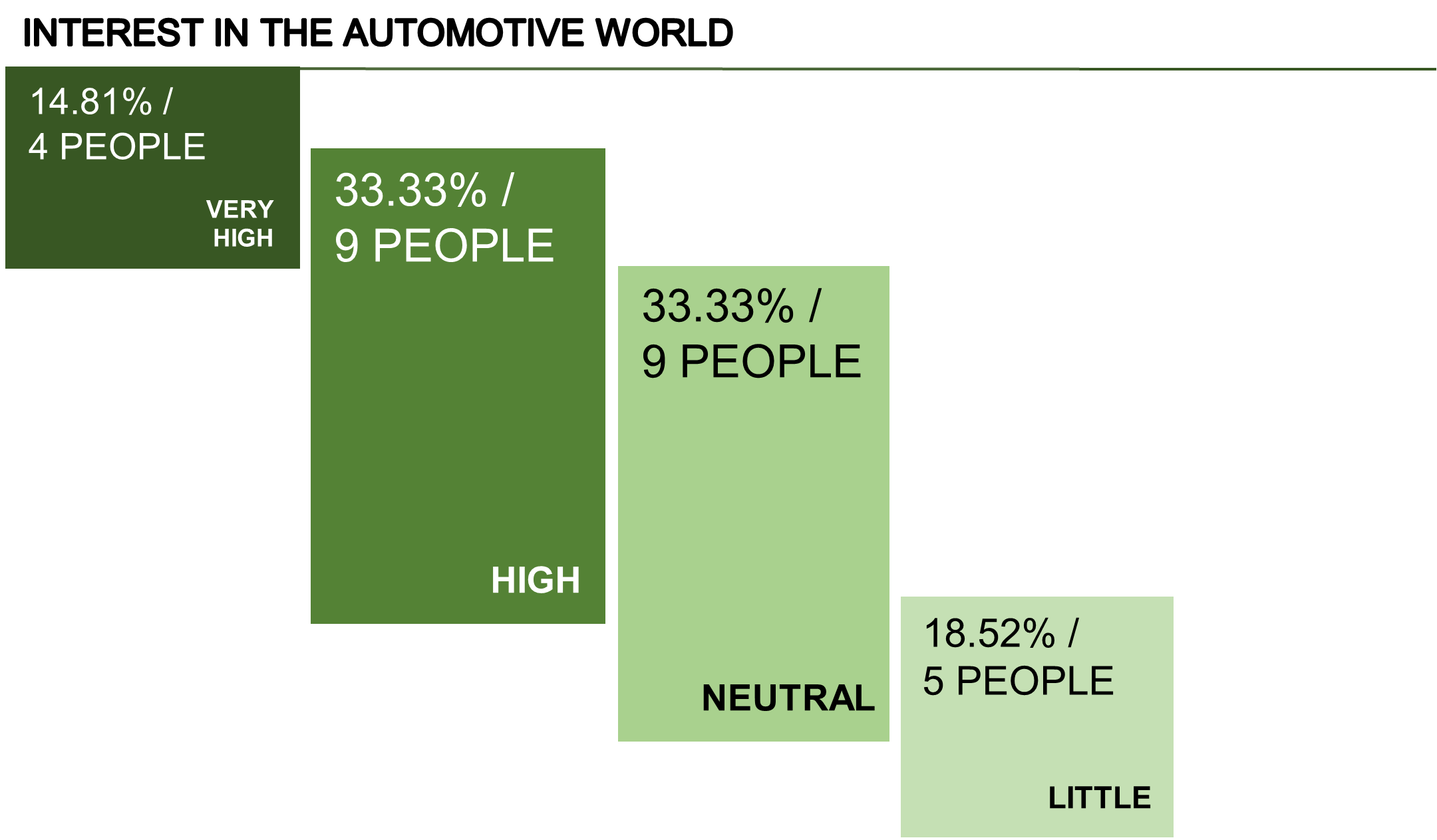}
         \label{fig:Interest}
     \end{subfigure}
     \begin{subfigure}[b]{0.98\linewidth}
        \centering
         \includegraphics[width=\linewidth]{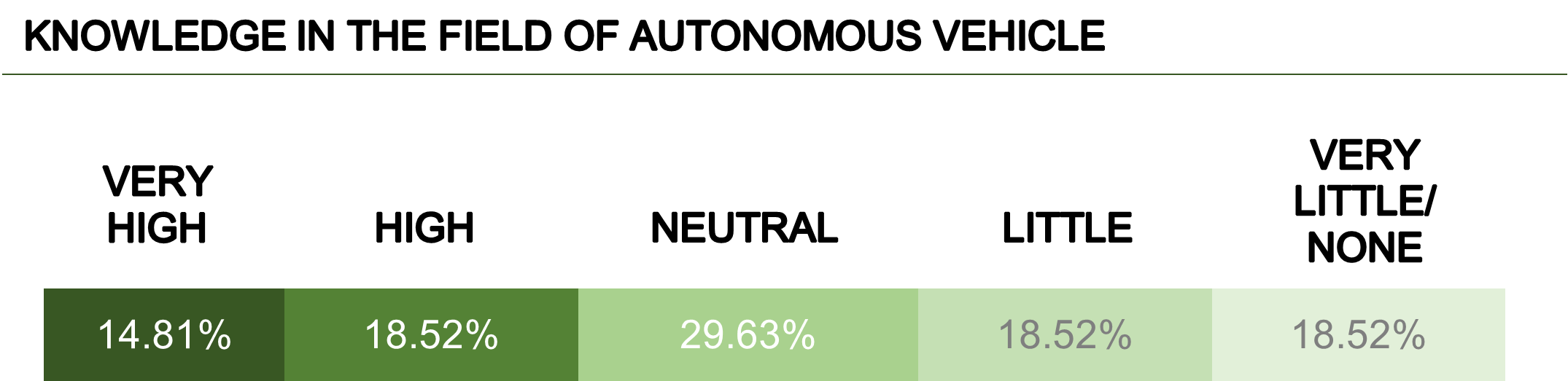}
         \label{fig:Knowledege}
     \end{subfigure}
        \caption{Statistics on knowledge in the topic}
        \label{fig:KnowledgeStats}
\end{figure}

Among the data collected in Figure~\ref{fig:KnowledgeStats}, one can observe the level of participant's interest in both the autonomous vehicles field and the automotive world. 48.14\% of the respondents expressed that their interest in the automotive world is high or very high. However, only 33\% of respondents had high or very high knowledge about autonomous vehicles. On the other hand, 30\% said that their knowledge in this field is neutral.

\begin{figure}[htbp]
\centerline{\includegraphics[width=\linewidth]{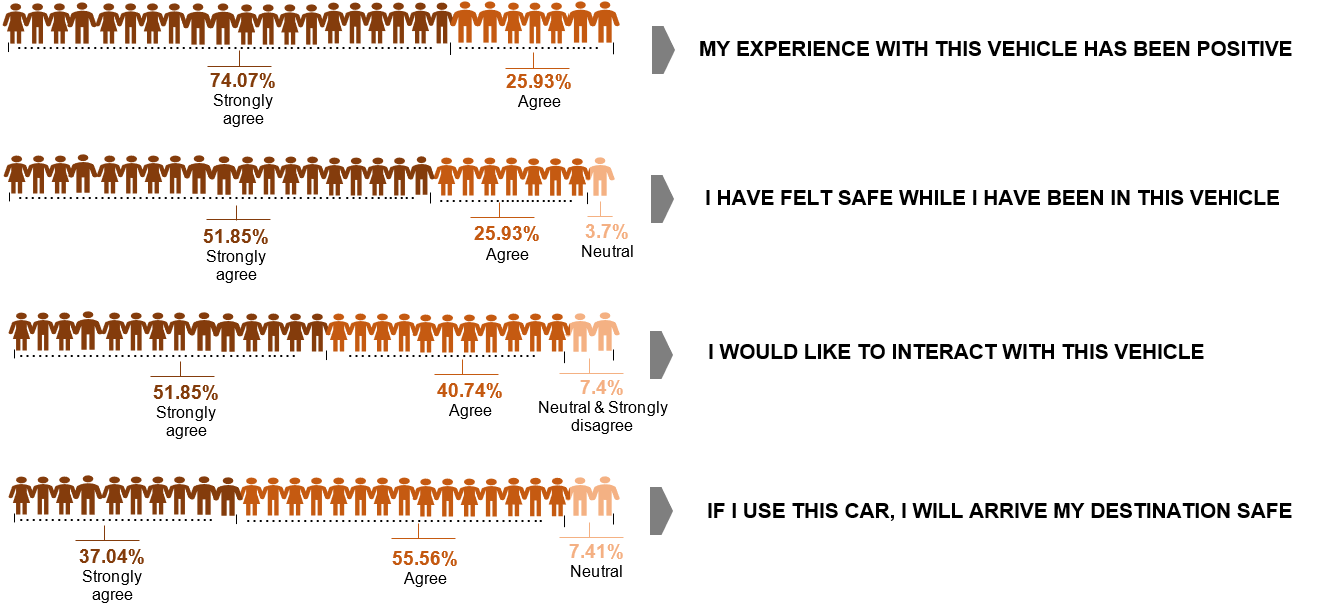}}
\caption{Statistics on use experience of the AVs.}
\label{fig:UseStats}
\end{figure}

One out of five candidates (18.52\%) expressed that their level of interest in this field is small, but twice as much people (37\%) expressed having little or very little knowledge in the self-driving car field. Finally, there were no specific age and gender differences among people who expressed high and very high interest in the autonomous vehicle field. The same was true for those with either neutral or little interest/knowledge in the automotive field. 

\begin{figure}[htbp]
\centerline{\includegraphics[width=\linewidth]{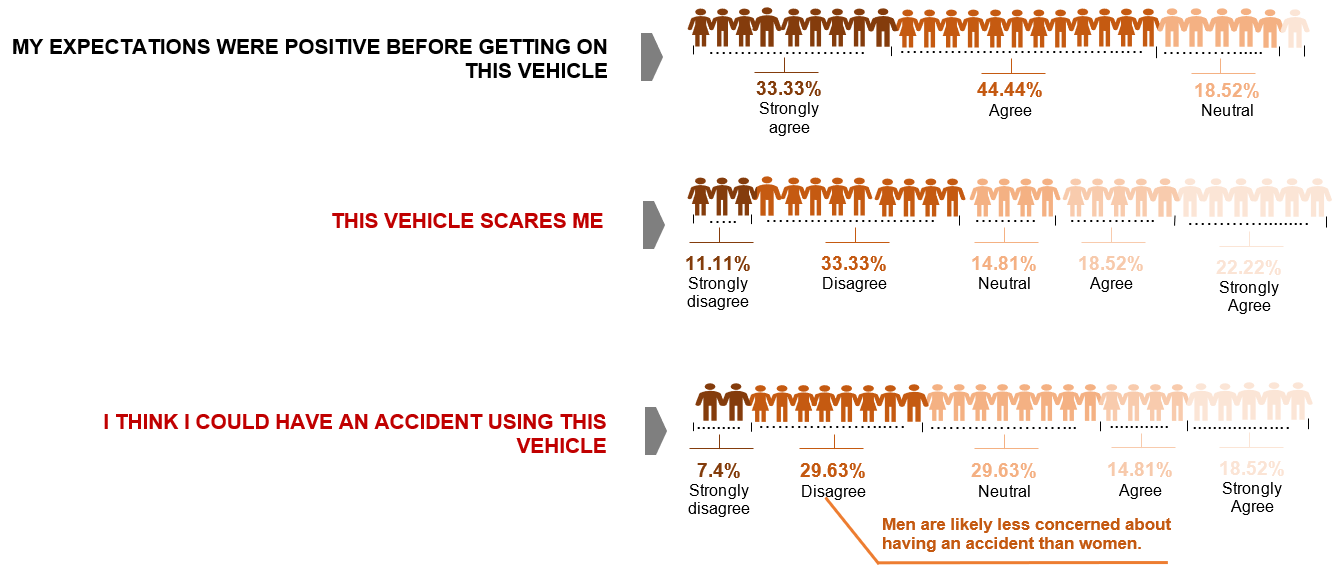}}
\caption{Complementary statistics about the use of AVs.}
\label{fig:UseStats2}
\end{figure}

As figures~\ref{fig:UseStats} and~\ref{fig:UseStats2} show, even though not all participants (78\%) had positive expectations before getting into the autonomous vehicle, interestingly, after the experiment, all people (100\%) reported that their experience with the car had been positive. Furthermore, 25 out of 27 people wanted to interact with the vehicle, and the same number of candidates assumed that they would arrive safely at their destination. 

Surprisingly, 26 people (96.3\%) of those participants stated that they felt safe in the vehicle; 11 people (41\%), however, said that this vehicle scared them, whereas almost the same number of people, 12 people (44\%) were not frightened by the car. Additionally, the candidate's other concern about using an autonomous vehicle was having an accident. One-third of people thought that they could have an accident using this vehicle. Yet, the same number of people (30\%) responded this question neutrally or disagreed. 

\begin{figure}[htbp]
\centerline{\includegraphics[width=\linewidth]{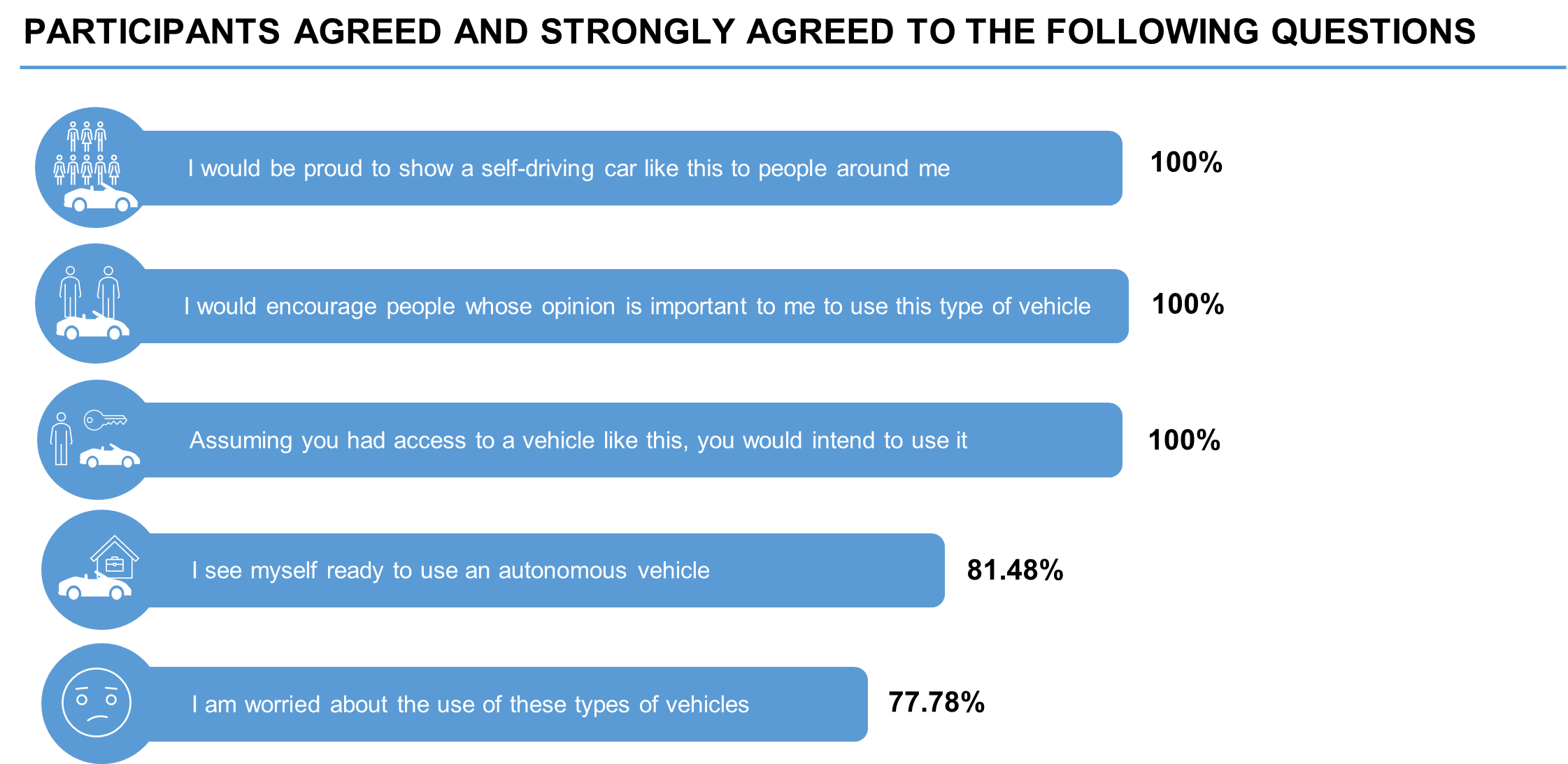}}
\caption{Statistics about willingness to use an AV.}
\label{fig:StatsWillingness}
\end{figure}

A high willingness for autonomous vehicle usage was observed among participants. As it can be seen in Figure~\ref{fig:StatsWillingness}, all participants would be proud to show a self-driving car to people around them and also encourage people whose opinion is important to them to use this type of vehicle. Even though 81.48~\% of candidates answered that they were ready to use an autonomous vehicle, surprisingly, all participants would intend to use an autonomous vehicle, if they had access to one. Interestingly, most people responded that they were worried about using this type of vehicle (77.8~\%). Clarifying users' reasons for concern would help discover further developments in the autonomous vehicle field. 

\section{Conclusions and Future Work}
\label{secIV:Conclusions}

This work comprises the demonstration of an SAE level 4 autonomous vehicle, in which a range of people have been able to participate. The objective of this demonstration was to bring this technology closer to society as well as to evaluate its impact and the degree of knowledge and acceptance of autonomous vehicles among the participants. To this end, all those who have had the opportunity of taking a ride aboard the ATLAS platform have been surveyed. 

As for the results presented in this paper, we consider as the most important finding that most participants found the experience of riding in an experimental autonomous car positive. In addition, the willingness to use such a car was extremely high. Furthermore, knowledge about the field of autonomous vehicles and interest in the automotive world seemed generally acceptable: almost half of the candidates (48.12\%) reported a high interest in the automotive world, and one-third of the respondents claimed to have high or very high knowledge in this field. However, a surprisingly high percentage of respondents expressed concern about the use of an autonomous vehicles (78\%), one-third of people were worried about getting into an accident (33\%), and 40.74\% of people were fearful of self-driving cars. 

Further exploration of these concerns would help to uncover future development opportunities in this field. To this end, questions such as those listed below could enhance the positive impact of autonomous vehicles on society. 
\begin{itemize}
    \item Why do people worry about using an AV?
    \item Why are people afraid of AVs?
    \item Why do people think they could get into an accident using an AV? 
\end{itemize}

Finding the answers to such questions and introducing improvements in ATLAS that help reduce the rejection of this technology and increase users' confidence are the actions to be taken in future work. 

\section*{Acknowledgment}

The authors want to acknowledge the work and support of the members of the Autonomous Mobility and Perception Lab (AMPL) that helped and assisted in the public demonstration: Sergio Campos Novoa, Carmen Barbero Ruiz, Diego Martin Maeso, David Mataix Borrell, Victoria Frutos Navarro and Borja Perez Frutos.

\vspace{12pt}

\end{document}